\documentclass[runningheads]{llncs}
\usepackage[T1]{fontenc}
\usepackage{graphicx}
\usepackage{amsmath}
\usepackage{amssymb}
\usepackage{booktabs}
\usepackage{microtype}
\usepackage{hyperref}
\usepackage{url}
\usepackage{makecell}
\usepackage{enumitem}
\usepackage{multirow}

\begin{document}
\title{IMFACT: Counterfactual Explanations for Time Series via Intrinsic Mode Function Substitution}
\titlerunning{Counterfactual Explanations via Intrinsic Mode Function Substitution}
%
\author{Udo Schlegel\inst{1,2} \and
Julian Rakuschek\inst{3} \and
Thomas Seidl\inst{1,2} \and\\
Andreas Holzinger\inst{4} \and
Tobias Schreck\inst{3} \and
Javier {Del Ser}\inst{5,6}}
\authorrunning{U. Schlegel et al.}
%
\institute{LMU Munich, Munich, Germany, \email{udo.schlegel@lmu.de}
\and
Munich Center for Machine Learning (MCML), Munich, Germany
\and
TU Graz, Institute of Visual Computing, Graz, Austria
\and
BOKU University Vienna, Vienna, Austria
\and
TECNALIA, Basque Research \& Technology Alliance (BRTA), Derio, Spain
\and
University of the Basque Country (EHU), Leioa, Spain
}
\maketitle              
\begin{abstract}
Oscillatory signals, such as vibration, carry class-discriminative information in specific frequency bands; perturbing them in raw feature space for counterfactual analysis easily destroys their temporal structure and produces physically implausible results. 
In this work, we introduce IMFACT (IMF-based counterfACTuals), a model-agnostic framework for generating plausible counterfactual explanations for time series classifiers that operates in the decomposition space of Empirical Mode Decomposition. 
An input signal is split into Intrinsic Mode Functions (IMFs), and selected IMFs are progressively substituted with those of a Nearest Unlike Neighbour (NUN) until the classifier flips to the target class. 
We evaluate six IMF-selection strategies and a multi-NUN cycling extension on two UCR benchmarks (FaultDetectionA, FruitFlies). 
The variance-based strategy with three NUNs outperforms two prominent baseline techniques on reliability and plausibility metrics, while cycling across three NUNs yields the best proximity across both datasets.

\keywords{Explainable AI \and Counterfactual Explanations \and Time Series Classification \and Empirical Mode Decomposition.}
\end{abstract}
%
%
%
\section{Introduction}
\label{sec:intro}

Explainability is a prerequisite for the responsible deployment of machine learning in high-stakes settings such as predictive maintenance, medical diagnosis, and structural health monitoring~\cite{rudin_stop_2019}. 
Explanation methods are divided into ante-hoc approaches, which build interpretability directly into the model, and post-hoc approaches, which explain a trained black-box model after the fact~\cite{guidotti_survey_2018,theissler_explainable_2022}. 
Among post-hoc explanation methods, counterfactual explanations hold a privileged position: they are contrastive and directly answer ``what minimal change to this input would have produced a different outcome?''~\cite{wachter_counterfactual_2017}, supporting actionable decision-making~\cite{chukwu_counterfactual_2025}. 
Oscillatory signals such as mechanical vibrations are particularly challenging in this regard, as their class-discriminative information is encoded in frequency bands rather than individual sample points~\cite{lei_review_2013}.

For tabular data, Wachter et al.~\cite{wachter_counterfactual_2017} formulated counterfactual generation as a constrained optimization problem that minimizes the distance to the decision boundary. 
For time series, Native Guide~\cite{delaney_instance_2021} improved upon this by anchoring perturbations to a Nearest Unlike Neighbor~(NUN), the closest training sample of a different class, providing an inductive bias towards in-distribution counterfactuals. 
Glacier~\cite{wang_glacier_2024} extended this by optimizing counterfactuals in a learned latent space regularized towards a NUN, improving plausibility over direct feature-space perturbation.
Despite these advances, existing methods often operate directly in raw feature space, which can easily produce implausible waveforms that violate frequency constraints, alter structure, or yield out-of-range amplitudes~\cite{schlegel_what_2026}.

In this paper, we propose IMFACT (IMF-based counterfACTuals), a model-agnostic framework that addresses the above limitations by operating in the decomposition space of Empirical Mode Decomposition~(EMD)~\cite{huang_empirical_1998}. 
EMD decomposes a signal into Intrinsic Mode Functions~(IMFs), amplitude- and frequency-modulated components without a predetermined basis. 
As a result, they have been extensively used for vibration-based fault detection because they carry physically interpretable frequency information~\cite{lei_review_2013,wu_ensemble_2009}. 
Our proposed IMFACT framework introduces a different perspective on counterfactual generation for time series by shifting the perturbation space from raw signal values to an EMD decomposition into IMFs. 
IMFACT progressively substitutes selected IMFs with those of a NUN until the classifier flips to the target class. 
Building on this idea, IMFACT defines IMF-selection strategies that explicitly control the trade-off between proximity and plausibility by targeting the signal's distinct frequency characteristics. 
By incorporating a multi-NUN extension, IMFACT enhances robustness and diversity in the search process by exploiting multiple target-class references. 
These design choices are validated on two datasets (FaultDetectionA and FruitFlies), considering six IMF-selection strategies and a multi-NUN cycling extension. 
Our results show that IMFACT achieves high validity while preserving temporal and amplitude properties, outperforming established baselines (Wachter~\cite{wachter_counterfactual_2017}, Native Guide~\cite{delaney_instance_2021}, and Glacier~\cite{wang_glacier_2024}) in validity and proximity metrics.

The remainder of this paper is structured as follows. 
\autoref{sec:background} reviews the background on counterfactual explanations, time series-specific approaches, and EMD. 
\autoref{sec:method} introduces the IMFACT framework and its main components. 
\autoref{sec:experiments} describes the experimental setup, including datasets, models, and evaluation metrics, while \autoref{sec:results} presents the results and comparative analysis. 
\autoref{sec:discussion} discusses the findings and, finally, \autoref{sec:conclusion} concludes the paper and outlines future research directions stemming from the limitations of our work.

\section{Background and Related Work}
\label{sec:background}

Before describing the proposed IMFACT framework, we review the foundations required by IMFACT, focusing on counterfactual explanations for time series (\autoref{ssec:CF_prob}), existing counterfactual generation approaches tailored to time series data (\autoref{ssec:TSCF}), and finally, Empirical Mode Decomposition and its relevance as a perturbation space (\autoref{ssec:EMD}).

\subsection{Counterfactual Explanations} \label{ssec:CF_prob}

A counterfactual explanation $\mathbf{x}'$ for an instance $\mathbf{x}$ with label $y = f(\mathbf{x})$ is a minimally modified version $\mathbf{x}'$ satisfying $f(\mathbf{x}') = y_{\mathrm{target}} \neq y$~\cite{wachter_counterfactual_2017}. 
Generation typically minimises:
\begin{equation}
  \mathcal{L}(\mathbf{x}') = \lambda \cdot d(\mathbf{x}, \mathbf{x}') + \ell\!\left(f(\mathbf{x}'), y_{\mathrm{target}}\right),
  \label{eq:cf_loss}
\end{equation}
where $d$ measures proximity and $\ell$ penalises target-class violation. Beyond \emph{validity} (the counterfactual is classified as the target class) and \emph{proximity} (the counterfactual is as close as possible to the original instance), the field converges on \emph{plausibility} (the counterfactual lies on the data manifold and resembles realistic instances), \emph{sparsity} (as few features as possible are changed to produce the class flip), and \emph{actionability} (the required changes are feasible for the user)~\cite{guidotti_counterfactual_2022,verma_counterfactual_2024,chukwu_counterfactual_2025,schlegel_what_2026}.

\subsection{Counterfactual Explanations for Time Series} \label{ssec:TSCF}

Wachter et al.~\cite{wachter_counterfactual_2017} provide a model-agnostic baseline via gradient optimization from a random initialization, but are not designed for temporal data and tend to produce diffuse, globally distributed changes. 
Native Guide~\cite{delaney_instance_2021} retrieves the NUN, extracts a class-discriminative weight vector from class activation maps, and guides perturbation between the query and the NUN, yielding temporally coherent, instance-grounded counterfactuals. 
Glacier~\cite{wang_glacier_2024} uses gradient search with local latent-space constraints; MASCOTS~\cite{pludowski_mascots_2025} operates in a symbolic feature space; Schlegel et al.~\cite{schlegel_interactive_2024} provide a human-in-the-loop method.
Schlegel and Seidl~\cite{schlegel_what_2026} provide a survey and taxonomy of counterfactual explanation methods for time series, organizing existing work into optimization-based, evolutionary, instance-based, latent-space, segment-based, and hybrid approaches, and highlighting temporal coherence, plausibility, and actionability as the core open challenges. 
However, many of these methods, e.g., MASCOTS cannot handle long time series instances to generate counterfactual in a fast fashion.

Despite this breadth of methods, none explicitly exploit signal decomposition as a perturbation space, a gap that IMFACT addresses by operating directly on the physically interpretable IMFs produced by EMD.

\subsection{Empirical Mode Decomposition} \label{ssec:EMD}

EMD~\cite{huang_empirical_1998} decomposes a signal $x(t)$ into IMFs $c_1(t), \ldots, c_K(t)$ and residual $r(t)$:
\begin{equation}
  x(t) = \sum_{k=1}^{K} c_k(t) + r(t).
  \label{eq:emd}
\end{equation}

The sifting algorithm iteratively identifies local extrema, interpolates upper and lower envelopes, subtracts their mean, and repeats until the IMF stopping criterion is met~\cite{luukko_introducing_2016}. IMFs are ordered from highest to lowest instantaneous frequency, forming a data-adaptive, complete, near-orthogonal basis suited to non-stationary and non-linear signals~\cite{huang_empirical_1998}.

\section{The IMFACT Framework}
\label{sec:method}

\begin{figure}[ht]
  \centering
  \includegraphics[width=\linewidth,trim={0 4.8cm 0 0},clip]{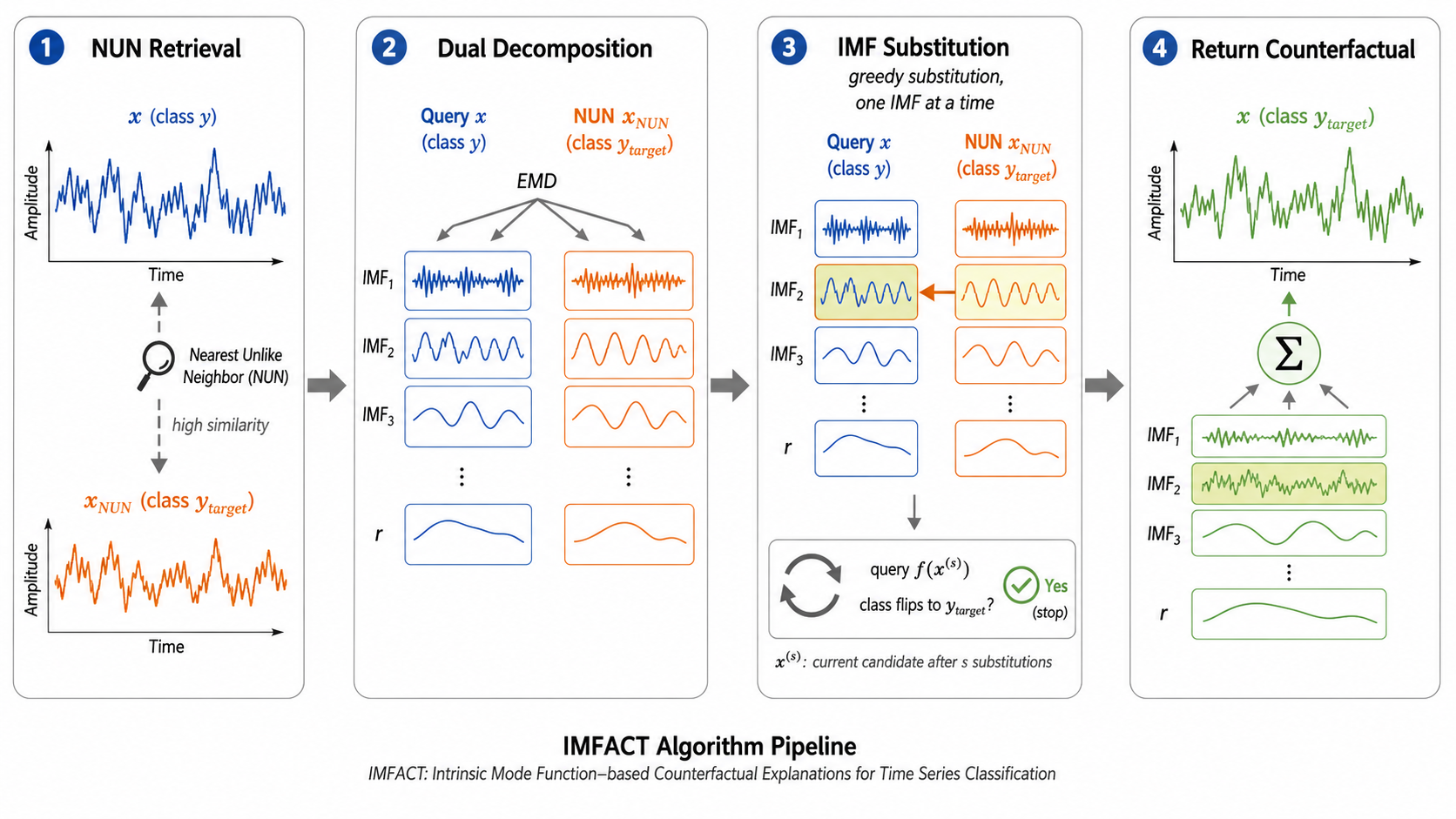}
  \caption{Overview of the IMFACT framework. \textbf{(1) NUN Retrieval:} the Nearest Unlike Neighbor $\mathbf{x}_{\mathrm{NUN}}$ of the target class $y_{\mathrm{target}}$ is retrieved for query $\mathbf{x}$ of class $y$. \textbf{(2) Dual Decomposition:} both signals are decomposed via EMD into IMFs $\mathrm{IMF}_1,\dots,\mathrm{IMF}_r$ plus a residual $r$, ordered from highest to lowest frequency. \textbf{(3) IMF Substitution:} IMFs are substituted from $\mathbf{x}_{\mathrm{NUN}}$ one at a time, in an order set by the chosen selection strategy (e.g.\ \texttt{distance}, \texttt{variance}, \texttt{extremes}), querying the classifier $f$ after each step until the prediction flips to $y_{\mathrm{target}}$. \textbf{(4) Return Counterfactual:} the IMFs are summed to give the final counterfactual $\mathbf{x}'$, preserving unsubstituted structure while incorporating only the minimal components needed to cross the decision boundary.}
  \label{fig:overview}  
\end{figure}

IMFACT generates counterfactual explanations by operating in the IMF decomposition space rather than directly in the raw time series domain. The core idea is to decompose both the query instance and a target-class reference into their constituent frequency modes, then iteratively substitute modes from the reference into the query until the classifier changes its prediction. 
\autoref{fig:overview} provides an overview of the pipeline. We first formalise the problem, then describe the four algorithmic stages, the IMF-selection strategies, and the multi-NUN extension.

\subsection{Problem Formulation}

Let $\mathbf{x} \in \mathbb{R}^T$ be a univariate time series of length $T$, classified by a black-box model $f : \mathbb{R}^T \to \mathcal{Y}$ as class $y = f(\mathbf{x})$. Given a target class $y_{\mathrm{target}} \neq y$, IMFACT aims to find a counterfactual $\mathbf{x}' \in \mathbb{R}^T$ that satisfies three criteria: 
\emph{validity} (the counterfactual is classified as the target class, i.e., $f(\mathbf{x}') = y_{\mathrm{target}}$); 
\emph{proximity} (the counterfactual stays as close as possible to the original instance, typically measured by $|\mathbf{x}' - \mathbf{x}|^2$); and 
\emph{range validity} (each time step of $\mathbf{x}'$ remains within the value range observed for $y_{\mathrm{target}}$ in the training data, ensuring basic plausibility).

\subsection{Algorithm Overview}

IMFACT proceeds in four stages, illustrated in \autoref{fig:overview}. The key intuition is that instead of perturbing the raw time series directly, we operate on its frequency decomposition, replacing one IMF at a time with the corresponding mode from a target-class reference until the decision boundary is crossed.

\vspace{2mm}
\noindent\textbf{Stage~1 --- NUN Retrieval.}
Given a query $\mathbf{x}$ with class $y$, we retrieve its Nearest Unlike Neighbor $\mathbf{x}_{\mathrm{NUN}}$ from the training set via nearest-neighbor search with Euclidean distance restricted to instances with label $y_{\mathrm{target}}$~\cite{delaney_instance_2021}. 
The NUN serves as a real, in-distribution representative of the target class and anchors the subsequent perturbation to the data manifold. 
For clarity, we describe the single-NUN case here; \autoref{sec:multi_nun} extends this stage to a pool of $n$ NUNs, from which IMFs are drawn according to a cycling or closest-neighbor policy.

\vspace{2mm}
\noindent\textbf{Stage~2 --- Dual Decomposition.} 
We apply EMD independently to both $\mathbf{x}$ and $\mathbf{x}_{\mathrm{NUN}}$, yielding IMF sets $\{c_k\}_{k=1}^K$ and $\{c_k^{\mathrm{NUN}}\}_{k=1}^{K'}$ together with residuals $r$ and $r^{\mathrm{NUN}}$, indexed from $k=1$ (highest-frequency oscillation) to $k=K$ (lowest-frequency oscillation), with $r$ capturing the remaining non-oscillatory trend. 
If the two decompositions produce a different number of IMFs ($K \neq K'$), we zero-pad the shorter set so that every index $k$ up to $\max(K, K')$ has a defined component in both signals, allowing IMFs to be compared and substituted position-by-position. 
Each IMF pair $(c_k, c_k^{\mathrm{NUN}})$ thus represents the same oscillatory scale in the query and reference signal, respectively, and only the query's residual $r$ is retained throughout, since it encodes the coarse trend that is not the target of frequency-band substitution.

\vspace{2mm}
\noindent\textbf{Stage~3 --- IMF Selection and Perturbation.} 
We select an ordered sequence of IMF indices $\mathcal{S} = (k_1, k_2, \ldots)$ according to the chosen strategy (\autoref{sec:strategies}). 
At each step $s$, we replace $c_{k_s}$ with $c_{k_s}^{\mathrm{NUN}}$ and iteratively reconstruct the candidate counterfactual as:
\begin{equation}
  \mathbf{x}^{(s)} = \sum_{k \notin \mathcal{S}_s} c_k \;+\; \sum_{k \in \mathcal{S}_s} c_k^{\mathrm{NUN}} \;+\; r,
  \label{eq:reconstruct}
\end{equation}
where $\mathcal{S}_s = \{k_1, \ldots, k_s\}$ is the set of already-substituted IMF indices, and $r$ is the query's own residual from Stage~2, kept fixed and never substituted with $r^{\mathrm{NUN}}$. 
We query the black-box classifier $f(\mathbf{x}^{(s)})$ after each substitution and stop as soon as $f(\mathbf{x}^{(s^*)}) = y_{\mathrm{target}}$. 
This greedy one-at-a-time substitution of IMFs ensures that the minimum number of IMFs is modified before validity is achieved, naturally favoring sparse and proximal counterfactuals. 
In the worst case, the loop runs for $|\mathcal{S}|$ steps, so the total query budget scales linearly in the number of IMFs rather than with the length $T$ of the raw series, which keeps the search efficient even for long signals.

\vspace{2mm}
\noindent\textbf{Stage~4 --- Return.} 
We return $\mathbf{x}' = \mathbf{x}^{(s^*)}$ as the final counterfactual explanation. 
If no valid counterfactual is found after all IMFs have been substituted, the method is considered unsuccessful for that instance; in this case $\mathbf{x}^{(s^*)}$ would coincide with $\mathbf{x}_{\mathrm{NUN}}$ itself (up to the retained residual $r$), and we report the instance as a failure rather than returning this fallback as an explanation.

\subsection{IMF-Selection Strategies}
\label{sec:strategies}

IMFACT relies on an ordering or weighting over IMFs that determines which modes are substituted first. Different strategies encode different hypotheses about where the class-discriminative information resides. We consider six strategies:
\begin{itemize}[leftmargin=*]
\item \texttt{distance}: For each IMF, we compute the Jensen--Shannon divergence between the power spectral density (PSD) of the current IMF and the PSD of the corresponding NUN IMF; IMFs with larger divergence are perturbed earlier. 
This favours modes whose frequency differs most between query and NUN.


\item \texttt{variance}: For each IMF index, we measure the difference in class-level variance between the source class and the target class, and normalise across IMFs. 
IMFs whose variance differs most between classes are perturbed first, under the assumption that these modes carry the strongest class-discriminative signal.

\item \texttt{extremes}: In this case, IMFACT computes the same distance as in \texttt{distance}, and then keep only the strongest and weakest IMF distances per channel, setting all other weights to zero. This focuses perturbation on a single most-different IMF and a single least-different IMF, yielding a highly sparse selection.


\item \texttt{coarse\_to\_fine}: We unlock IMFs gradually from coarse to fine scales over iterations. 
At early iterations only the coarsest (lowest-frequency) IMF is eligible for substitution. 
Every \texttt{$s_{\text{coarse}}$} iterations one additional, finer IMF becomes active. Within the active set, distances are computed as in \texttt{distance} and normalised across IMFs.
\end{itemize}

\subsection{Multi-NUN Extension}

The basic IMFACT framework uses a single NUN as the target-class reference. To reduce dependence on any single neighbour and to enrich the pool of candidate IMFs, we introduce a multi-NUN extension that retrieves the $n$ NUNs $\{\mathbf{x}_{\mathrm{NUN}}^{(i)}\}_{i=1}^n$ from the training set. At each substitution step $s$, IMFACT chooses from this set according to one of two policies:
\begin{itemize}[leftmargin=*]
\item \textbf{cycle}: IMFACT cycles through the $n$ NUNs in round-robin order. At step $s$, the next IMF is taken from NUN $i = (s \bmod n)$, ensuring that all NUNs contribute IMFs through the counterfactual search rather than relying on a single reference.

\item \textbf{closest}: At each step, for the IMF index $k$ selected by the current strategy, IMFACT chooses the NUN whose corresponding IMF $c_k^{(i)}$ is closest in L2 distance to the query IMF $c_k$. This greedily selects the most similar target-class mode for substitution at that scale.
\end{itemize}

\section{Experimental Setup} \label{sec:experiments}

In this section, we describe the experimental protocol used to evaluate IMFACT. 
We first outline the benchmark datasets (\autoref{ssec:datasets}) and classifier architecture (\autoref{ssec:classifier}), then detail the counterfactual evaluation metrics (\autoref{ssec:metrics}) and baseline methods (\autoref{ssec:baselines}), which are tested against IMFACT. 

\subsection{Datasets} \label{ssec:datasets}

Our experiments use two benchmark datasets from the Time Series Classification (TSC) website~\cite{middlehurst_aeon_2024} (which hosts and standardizes the UCR archive~\cite{dau_ucr_2019}) to ensure comparability with prior work on time series counterfactual explanations:
\begin{itemize}[leftmargin=*]
\item \textbf{FaultDetectionA}~\cite{lessmeier_condition_2016} contains one-dimensional accelerometer recordings from an electromechanical drive system with three operating conditions (healthy and two fault classes). 
The signals are recorded under controlled laboratory conditions and capture characteristic bearing and shaft vibration patterns that change when faults occur, making this dataset representative of the predictive-maintenance setting in which EMD has been widely applied~\cite{lei_review_2013,wu_ensemble_2009}. 

\item \textbf{FruitFlies} consists of wing-beat recordings from different \emph{Drosophila} species, where each univariate time series encodes an individual fly’s oscillatory wing-beat pattern over time. 
The task is to distinguish species based on subtle differences in frequency content and waveform shape, providing a complementary biological benchmark to the mechanical signals of FaultDetectionA. 
\end{itemize}

We use the standard train--test split defined on the TSC website, treating each univariate time series as a single-channel input to the classifier.
Both datasets exhibit non-stationary, oscillatory behavior, with class-discriminative information concentrated in specific frequency bands rather than at individual time points, which makes them suited for evaluating IMF-based counterfactuals built on EMD.

\subsection{Classifier} \label{ssec:classifier}

We use a lightweight Convolutional Neural Network (SimpleCNN) implemented in PyTorch~\cite{paszke_pytorch_2019} as the black-box classifier across all experiments, following Schlegel and Seidl~\cite{schlegel_what_2026}. 
The architecture comprises four convolutional blocks, each consisting of a \texttt{Conv1d} layer with a stride of 2, batch normalization, ReLU activation, and dropout, followed by two fully-connected layers with batch normalization and dropout, and a softmax output layer. 
The four convolutional layers use 16, 32, 64, and 128 channels, respectively, with kernel sizes of 5, 5, 3, and 3, progressively downsampling the input; the fully connected layers reduce the flattened representation to 256 and then to 128 units before the final classification head. 
The model takes a single-channel time series of length $T$ as input and outputs the softmax probabilities over the target classes. 
The model achieves an F1 score of 0.99 on the FaultDetectionA test dataset and 0.87 on the FruitFlies test dataset.
Since IMFACT is model-agnostic and only requires black-box query access, the choice of classifier does not affect the counterfactual generation procedure itself.










\subsection{Evaluation Metrics} \label{ssec:metrics}
We report a mixture of standard counterfactual metrics and aggregate scores inspired by a Keane-style evaluation~\cite{keane_if_2021}, computed from per-instance quantities and then averaged over the evaluated set. 
Specifically, we measure \textbf{Validity}, the proportion of $\mathbf{x}$ for which one valid counterfactual $\mathbf{x}'$ is found; \textbf{Average L2}, the average Euclidean distance $\|\mathbf{x}' - \mathbf{x}\|_2$ between sample and cf; \textbf{Pct. Changed}, the fraction of points that differ between $\mathbf{x}$ and $\mathbf{x}'$ (one minus sparsity); \textbf{Range Val.}, the fraction of time steps in $\mathbf{x}'$ that remain within the empirical value range observed for the target class in the training data; \textbf{Autocorr.}, the average Pearson correlation between the lag-1 autocorrelation profiles of $\mathbf{x}$ and $\mathbf{x}'$; and \textbf{Time}, the average time needed to generate a counterfactual for a single query.

\subsection{Baselines} \label{ssec:baselines}

We compare IMFACT against two widely used counterfactual baselines and one more recent approach, which is suitable for long time series data:
\begin{itemize}[leftmargin=*]
    \item \textbf{Wachter}~\cite{wachter_counterfactual_2017}: 
    A model-agnostic optimization-based method originally proposed for tabular data. It formulates counterfactual generation as minimizing a loss that trades off proximity to the original instance against a loss enforcing the desired target class, without making any time-series-specific assumptions.

    \item \textbf{Native Guide}~\cite{delaney_instance_2021}: 
    An instance-based method specifically designed for time series. Native Guide retrieves a Nearest Unlike Neighbor from the training set and uses a class-discriminative weight vector (e.g., derived from class activation maps) to highlight and perturb the most important regions of the series towards the guide, thereby producing sparse, proximal counterfactuals.

    \item \textbf{Glacier}~\cite{wang_glacier_2024}: 
    A latent-space counterfactual method for time series. Glacier performs locally constrained optimization over an autoencoder's latent space, explicitly enforcing smoothness and neighborhood consistency, guided by saliency constraints that localize perturbations to yield more targeted edits.
\end{itemize}

For a fair comparison, all four methods (IMFACT, Native Guide, Wachter, Glacier) are evaluated on the same 50 randomly selected test instances per dataset using the implementation of Schlegel and Seidl~\cite{schlegel_what_2026}.
Experiment results and source code can be found online with more in-depth analysis~\footnote{
Code and results can be found in the \href{https://github.com/visual-xai-for-time-series/counterfactual-explanations-for-time-series/}{CFTS Github Repo} at \href{https://github.com/visual-xai-for-time-series/counterfactual-explanations-for-time-series/tree/main/cfts/cf_imfact}{cf\_imfact}.}.

\section{Results}
\label{sec:results}

We structure the evaluation in two parts. First, we present a combined ablation across the IMF-selection strategy and the multi-NUN configuration, on both datasets, to assess how both design choices jointly affect proximity, plausibility, and runtime. 
Second, we compare the best-performing IMFACT configuration against Native Guide, Wachter, and Glacier on both datasets to situate IMFACT within the broader landscape of time series counterfactual methods.

\subsection{IMF-Strategy and Multi-NUN Ablation}
\label{sec:strategy_nun_ablation}

We ablate two design choices jointly: the IMF-selection strategy (\texttt{distance}, \texttt{variance}, \texttt{extremes}, \texttt{coarse\_to\_fine}) and the multi-NUN configuration -- the number of neighbours ($n\!\in\!\{1,2,3,5\}$) and, for the \texttt{distance} strategy specifically, the neighbour-selection policy (\texttt{cycle} rotates round-robin through the NUN pool; \texttt{closest} greedily selects the locally nearest NUN at each step). 
\texttt{distance} is the only strategy for which a \texttt{closest}-policy sweep was run; all other strategy rows use \texttt{cycle}. 
The full grid comprises 19 configurations per dataset (38 total), all of which reach full success (100.0\%), so there is no failure case anywhere in this ablation. 
\autoref{tab:ablation_highlights} reports the best- and second-best-performing configurations by each metric, the fastest configuration, our recommended default, and a weakest-strategy reference point, on both datasets (Success and Pct.\ Changed are omitted, since every configuration reaches 100.0\% success and $\geq\!99.9$\% Pct.\ Changed); the full grid is provided in the project repository.\footnote{Ablation grid for the datsets: \url{https://github.com/visual-xai-for-time-series/counterfactual-explanations-for-time-series/tree/main/cfts/cf_imfact}}

\begin{table}[h!tb]
\caption{Highlights of the combined IMF-strategy $\times$ multi-NUN ablation. \texttt{distance (dist)}, \texttt{variance (var)}, \texttt{extremes (ext)}, \texttt{coarse\_to\_fine (ctf)}; \texttt{cycle (cyc)}, \texttt{closest (clst)}. Bold/underline mark the best/second-best value per column across the grid per dataset; the \texttt{var} reference row is unmarked. Full grid in the project repo.}
\label{tab:ablation_highlights}
\centering
\setlength{\tabcolsep}{2pt}
\renewcommand{\arraystretch}{1.05}
\resizebox{\columnwidth}{!}{\begin{tabular}{llp{5cm}cccc}
\toprule
Data & Config. & Highlight & \makecell[c]{Average\\L2} & \makecell[c]{Range\\Val.} & Autocorr. & Time (s) \\
\midrule
\multirow{8}{*}{\rotatebox[origin=c]{90}{FaultDetectionA}}
& ext\_n1\_cyc  & Best L2; best range validity          & \textbf{13.476} & \textbf{0.959} & 0.634 & 0.905 \\
& ext\_n3\_cyc  & 2nd-best L2; 2nd-best range validity  & \underline{13.952} & \underline{0.954} & 0.762 & 0.968 \\
& ctf\_n3\_cyc  & Best autocorrelation                  & 14.832 & 0.947 & \textbf{0.886} & 0.799 \\
& ctf\_n5\_cyc  & 2nd-best autocorrelation              & 14.941 & 0.947 & \underline{0.882} & 0.825 \\
& dist\_n1\_cyc & Fastest                               & 17.478 & 0.938 & 0.750 & \textbf{0.375} \\
& dist\_n3\_cyc & Recommended default                   & 16.241 & 0.943 & 0.802 & 0.415 \\
& dist\_n5\_clst & Best autocorr.\ under \texttt{clst}  & 17.929 & 0.934 & 0.804 & 0.455 \\
& var\_n1\_cyc  & Weakest strategy (reference)          & 25.494 & 0.903 & 0.805 & 11.022 \\
\midrule
\multirow{8}{*}{\rotatebox[origin=c]{90}{FruitFlies}}
& ext\_n3\_cyc  & Best L2                               & \textbf{0.650} & 0.994 & 0.989 & 0.795 \\
& ext\_n5\_cyc  & 2nd-best L2                           & \underline{0.679} & 0.994 & 0.992 & 0.814 \\
& ctf\_n3\_cyc  & Best range validity                   & 0.744 & \textbf{0.998} & 0.995 & 0.684 \\
& dist\_n3\_clst & 2nd-best range validity              & 0.698 & \underline{0.997} & 0.995 & 0.537 \\
& dist\_n2\_clst & Best autocorrelation                 & 0.781 & 0.992 & \textbf{0.997} & 0.569 \\
& dist\_n2\_cyc & 2nd-best autocorrelation              & 0.714 & 0.996 & \underline{0.996} & 0.514 \\
& dist\_n3\_cyc & Fastest; recommended default          & 0.696 & 0.996 & 0.994 & \textbf{0.449} \\
& var\_n3\_cyc  & Weakest strategy (reference)          & 1.121 & 0.985 & 0.988 & 5.494 \\
\bottomrule
\end{tabular}}
\renewcommand{\arraystretch}{1}
\end{table}

Reading across the two blocks separates the two ablation effects. 
\textbf{Strategy effect}: on FaultDetectionA, \texttt{extremes} gives the best proximity and range validity, while \texttt{coarse\_to\_fine} gives the best autocorrelation preservation; \texttt{variance} is the weakest strategy by a wide margin, way slower than any other strategy and worst on both L2 (25.494) and range validity (0.903) at its best-performing $n$. 
On FruitFlies, \texttt{coarse\_to\_fine} and \texttt{extremes} again lead on plausibility and proximity, and \texttt{variance} is again the slowest and least proximal strategy.

\textbf{Multi-NUN effect}: within the \texttt{distance} strategy,the only one with both $n$ and policy varied, \texttt{cycle} outperforms \texttt{closest} on proximity at every neighbour count on FaultDetectionA (e.g.\ 16.241 vs.\ 18.619 at three neighbours), while on FruitFlies the ordering reverses at five neighbours, where \texttt{closest} becomes the most proximal configuration in the entire \texttt{distance} block (0.687). 
\texttt{closest} does, however, attain the best autocorrelation preservation on FaultDetectionA at five neighbours (0.804, \texttt{multi\_nun\_closest\_n5}) and the outright best autocorrelation preservation on FruitFlies at two neighbours (0.997, \texttt{multi\_nun\_closest\_n2}), so the \texttt{closest} policy is not uniformly worse, it simply trades proximity for plausibility in most settings. 
We settled on \texttt{distance}, $n\!=\!3$, \texttt{cycle} as the recommended default because it is consistently near the top of the grid on both datasets without being a narrow single-metric winner: best-or-near-best range validity on FaultDetectionA (0.943) and within 0.009 of the best FruitFlies L2 (0.696 vs.\ 0.687), while avoiding the runtime volatility discussed next.

\emph{On runtime}: \texttt{distance} configurations are the fastest strategy on both datasets (0.375--0.455s on FaultDetectionA, 0.449--0.569s on FruitFlies, excluding one outlier discussed below). 
\texttt{variance} configurations are consistently and substantially the slowest (11.0--11.1s on FaultDetectionA, 5.5--5.6s on FruitFlies); \texttt{extremes} and \texttt{coarse\_to\_fine} sit in an intermediate band (roughly 0.76--1.04s on FaultDetectionA, 0.66--0.98s on FruitFlies) that was stable in the latest pass but showed order-of-magnitude spikes in earlier passes, so we would still recommend averaging over additional runs before treating any intermediate-band figure as final. 
One remaining single-point anomaly, visible in the full repository grid: \texttt{multi\_nun\_closest\_n5} on FruitFlies measured at 5.301s in the latest pass versus 0.53--0.61s previously and versus every neighbouring configuration in the same block, which we treat as noise rather than a genuine cost.

\subsection{Baseline Comparison}

\begin{figure}[h!b]
  \centering
  \includegraphics[width=\linewidth,trim=0 0 0 4cm,clip=true]{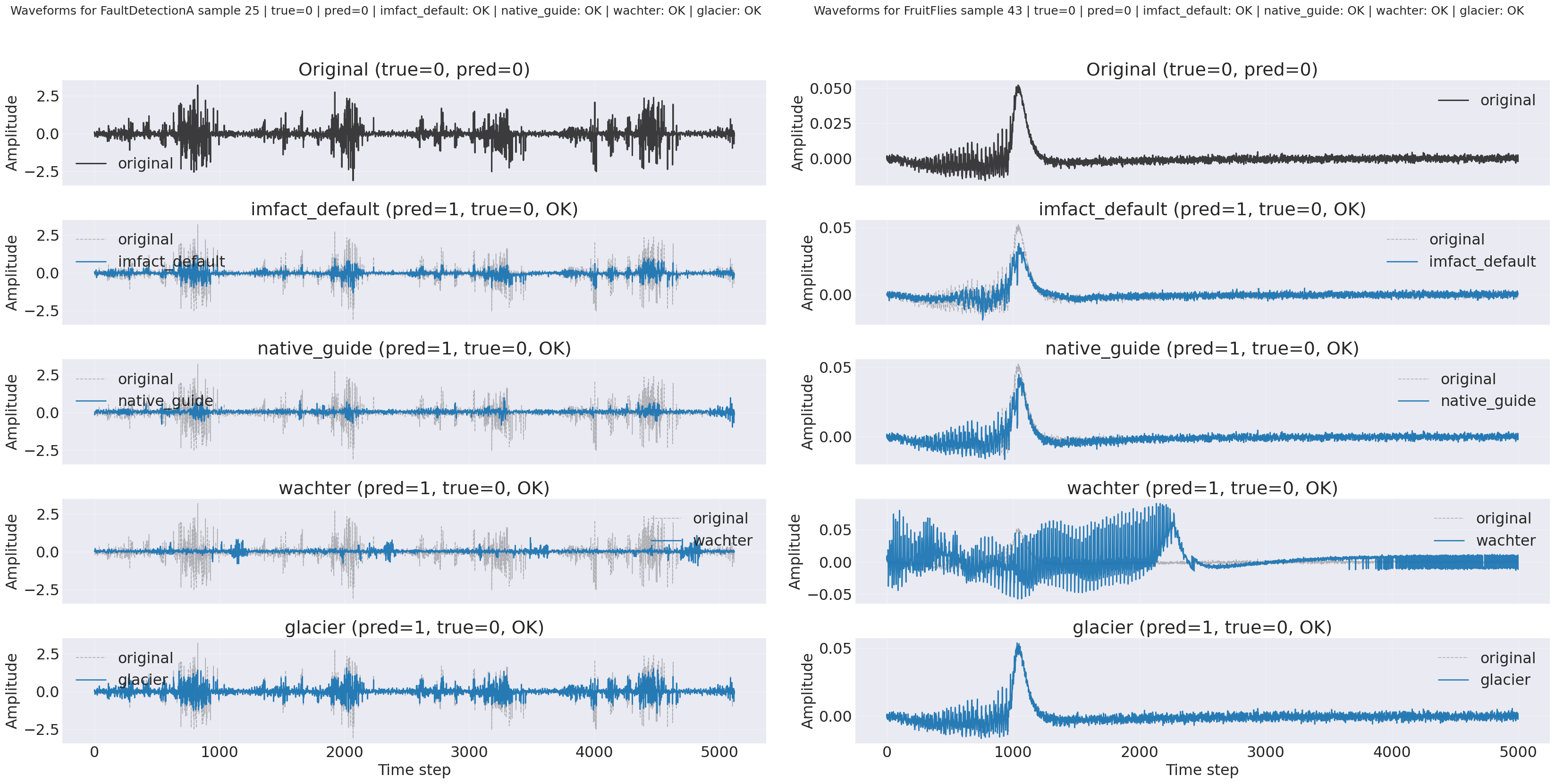}
  \caption{Counterfactual waveforms for representative instances on FaultDetectionA (sample~25, true class~0, left) and FruitFlies (sample~43, true class~0, right). All four methods (IMFACT, Native Guide, Wachter, and Glacier) reach a valid counterfactual on both instances. IMFACT, Native Guide, and Glacier track the original signal's morphology closely on both datasets, while Wachter produces a visibly more divergent waveform on both instances despite reaching validity, illustrating that validity alone does not guarantee visual proximity to the original signal.}
  \label{fig:qualitative}
\end{figure}

\autoref{tab:baseline} reports the direct comparison of IMFACT against Wachter~\cite{wachter_counterfactual_2017}, Native Guide~\cite{delaney_instance_2021}, and Glacier~\cite{wang_glacier_2024} on both datasets.
IMFACT is one of three methods to reach full validity on FaultDetectionA (100.0\%, tied with Glacier and Native Guide) and one of two to reach full validity on FruitFlies (100.0\%, tied with Native Guide); Wachter reaches only partial validity on both datasets (58.0\% and 30.0\%), and Glacier's validity collapses to 36.0\% on FruitFlies despite being fully valid on FaultDetectionA. 
\autoref{fig:qualitative} shows a representative successful instance from each dataset, illustrating how the four methods differ in the shape of the counterfactuals they produce even when all reach validity.

\begin{table}[htb]
\caption{Baseline comparison against Wachter~\cite{wachter_counterfactual_2017}, Native Guide~\cite{delaney_instance_2021}, and Glacier~\cite{wang_glacier_2024}. Bold/underline mark best/second-best per column within each dataset, excluding Time and Pct.\ Changed; Validity is unmarked wherever methods tie for best. FruitFlies figures are carried over from the previous measurement pass (\autoref{sec:strategy_nun_ablation}).}
\label{tab:baseline}
\centering
\setlength{\tabcolsep}{4.6pt}
\begin{tabular}{llcccccc}
\toprule
Data & Method & \makecell[c]{Validity\\(\%)} & \makecell[c]{Average\\L2} & \makecell[c]{Pct.\\Changed} & \makecell[c]{Range\\Val.} & Autocorr. & Time (s) \\
\midrule
\multirow{4}{*}{\rotatebox[origin=c]{90}{\scriptsize \makecell{Fault-\\DetectionA}}}
& Glacier      & 100.0 & \textbf{8.788}  & 99.961 & 0.936          & \textbf{0.960} & 1.327 \\
& \textbf{IMFACT} & 100.0 & 18.862       & 100.000 & \underline{0.939} & 0.685       & \textbf{0.388} \\
& Native Guide & 100.0 & 20.843          & 58.974 & 0.922          & \underline{0.837} & 7.218 \\
& Wachter      & 58.0  & \underline{14.751} & 99.997 & \textbf{0.947} & 0.825       & 11.792 \\
\midrule
\multirow{4}{*}{\rotatebox[origin=c]{90}{FruitFlies}}
& Glacier      & 36.0  & 3.550           & 99.984 & 0.198          & 0.174          & 19.575 \\
& \textbf{IMFACT} & 100.0 & \underline{0.922} & 99.976 & \textbf{0.984} & \underline{0.987} & \textbf{0.239} \\
& Native Guide & 100.0 & \textbf{0.871}  & 54.534 & \underline{0.980} & \textbf{0.989} & 111.910 \\
& Wachter      & 30.0  & 1.070           & 100.000 & 0.959          & 0.985          & 0.059 \\
\bottomrule
\end{tabular}
\end{table}

On FaultDetectionA, Native Guide~\cite{delaney_instance_2021} reaches full validity in aggregate, but has the lowest range validity of the four methods on this dataset (0.922) and, notably, changes only 58.974\% of timesteps on average (\autoref{tab:baseline}) versus ~100\% for every other method, confirming that its edits are concentrated in a small region rather than spread across the series. 
On FruitFlies, Native Guide also reaches full validity and, in fact, attains the lowest L2 distance of any method (0.871) and the highest autocorrelation preservation (0.989), narrowly ahead of IMFACT, despite changing only 54.534\% of timesteps on average, the lowest Pct.\ Changed of any method on either dataset. 
Its main drawback is computational cost: Native Guide is roughly $19\times$ slower than IMFACT on FaultDetectionA, a ratio that has held consistently across repeated measurement passes even as the absolute times drifted (8.901s vs.\ 0.471s in one pass, 7.218s vs.\ 0.388s in the latest). 
On FruitFlies, the available measurements disagree substantially on magnitude (Native Guide at $\sim\!43\times$ slower in one run and $\sim\!468\times$ slower, 111.910s vs.\ 0.239s, in another), so while Native Guide is consistently and substantially the slower method, we do not report a specific multiplier for FruitFlies pending a more stable runtime measurement.

Wachter et al.~\cite{wachter_counterfactual_2017} shows a different trade-off: on FaultDetectionA it achieves lower L2 (14.751) and higher range validity (0.947) than IMFACT whenever it succeeds, and it remains among the fastest methods on FruitFlies (0.059s), but it succeeds on only 58.0\% of FaultDetectionA instances and 30.0\% of FruitFlies instances. 
This confirms the well-known limitation that unconstrained gradient-free optimisation frequently fails to reach the decision boundary, even when the counterfactuals it does produce look reasonable in isolation~\cite{schlegel_what_2026}; \autoref{fig:qualitative} illustrates the correspondingly larger visual departure from the original signal that Wachter's successful counterfactuals still exhibit. 
Wachter's own runtime on FaultDetectionA has been especially unstable across measurement passes, rising from 3.888s to 11.792s, now the slowest of the four methods on this dataset, which we flag as a further symptom of the runtime measurement noise discussed above rather than a stable characterisation of the method.

Glacier~\cite{wang_glacier_2024} is the strongest method on FaultDetectionA by nearly every metric other than validity, where it ties IMFACT and Native Guide at 100.0\%: it attains the lowest L2 (8.788) and the highest autocorrelation preservation (0.960) of any method on this dataset. 
This strength does not transfer to FruitFlies, however, where Glacier's validity collapses to 36.0\% and its range validity and autocorrelation preservation both fall below 0.2 (0.198 and 0.174, respectively), indicating that a large fraction of its counterfactuals fall outside the plausible signal range entirely.

IMFACT is the only method that achieves full validity on both datasets while avoiding both of these failure modes: Glacier's dataset-dependent collapse and Native Guide's high computational cost. 
Its own proximity and plausibility metrics are not uniformly best, Glacier is more proximal and better at preserving autocorrelation on FaultDetectionA, and Native Guide is marginally more proximal on FruitFlies, but IMFACT is the only method to combine full validity, competitive plausibility (range validity of 0.939 and 0.984 on FaultDetectionA and FruitFlies, respectively), and low runtime (0.388s on FaultDetectionA; 0.239--0.271s across FruitFlies measurement passes) consistently across both settings. 
Notably, IMFACT has also been the most \emph{stable} method on runtime across repeated measurement: while Glacier, Native Guide, and Wachter have each shown multi-second to two-orders-of-magnitude swings between passes, IMFACT's runtime has stayed within a narrow band on both datasets.

\textbf{Summary --}
Each baseline exhibits a distinct failure mode that IMFACT avoids: Glacier's validity collapses on FruitFlies, Wachter reaches the decision boundary only partially on either dataset, and Native Guide matches IMFACT's validity but at far higher and less stable runtime. IMFACT is thus the only method combining validity, plausibility, and stable low runtime across datasets.

\section{Discussion}
\label{sec:discussion}

\hspace{\parindent}\textit{Why EMD works as a perturbation space --}
IMFs are physically interpretable and hierarchically ordered in terms of influence. 
In vibration signals, high-index IMFs capture bearing resonance and fault-characteristic frequencies, while low-index IMFs encode broadband noise~\cite{lei_review_2013,wu_ensemble_2009}.
In our ablation (\autoref{tab:ablation_highlights}), it is \texttt{extremes} rather than \texttt{variance} that yields the closest, most plausible counterfactuals, and \texttt{coarse\_to\_fine} that best preserves autocorrelation. 
This suggests the classifiers are more sensitive to localized amplitude extrema than to the single most energetically dominant mode: substituting that mode wholesale, as \texttt{variance} does, requires the fewest search iterations (12--20 vs.\ 61--103 for the other strategies) but each such step is far larger and costlier, leaving \texttt{variance} both the least proximal strategy and, by over $10\times$, the slowest. 
Reaching the decision boundary quickly along the variance axis evidently means overshooting it in signal space, though we infer this from aggregate ablations rather than from direct evidence that IMF tracks what the classifier actually uses; confirming it would need instance-level checks such as attributions to individual IMFs~\cite{schlegel_visual_2023}.


\textit{On evaluation and multi-metric validation --}
A single validity metric is insufficient for assessing counterfactual quality~\cite{delser_generating_2024}, because each baseline in our study looks strong on validity alone while failing along others. 
Glacier~\cite{wang_glacier_2024} reaches full validity and the best proximity and autocorrelation on FaultDetectionA, yet its validity collapses to 36.0\% on FruitFlies, where its remaining counterfactuals are also implausible (range validity and autocorrelation both $<0.2$); a validity score computed on a single dataset would miss this dataset-dependent collapse. 
Wachter et al.~\cite{wachter_counterfactual_2017} is only partially valid on both datasets (58.0\%, 30.0\%), so its seemingly competitive proximity is measured over an easier subset of instances rather than the full evaluation set. 
Native Guide~\cite{delaney_instance_2021} reaches full validity and even leads on FruitFlies proximity and autocorrelation, but does so by concentrating its edits in a small region of the series, changing only 54.5--59.0\% of timesteps versus roughly 100\% for the other methods; this is a plausibility problem that validity and proximity alone cannot reveal. 
Together these cases show that validity, proximity, and plausibility can each look good in isolation while masking a failure visible only in another metric, which is why we report them jointly, alongside runtime, rather than relying on any single score~\cite{kostrzewa_towards_2026,schlegel_what_2026,verma_counterfactual_2024}. 
For the same reason, we report runtime stability rather than a single measurement: several baselines' wall-clock times varied by an order of magnitude or more across repeated passes (\autoref{sec:strategy_nun_ablation}), so a one-shot timing figure would be misleading on its own.

\textit{Generality and transferability --}
EMD applies to any non-stationary oscillatory signal without domain-specific adaptation. 
The FruitFlies results confirm transfer to a biological signal with a different frequency profile: IMFACT is the only method to reach full validity on both datasets while remaining plausibility-competitive (range validity 0.939 and 0.984) and runtime-stable across measurement passes. 
The NUN-guided substitution is also classifier-agnostic, requiring no gradients or model internals, making it applicable to any black-box classifier, with natural extensions to electroencephalography and financial time series, which share the non-stationary character IMF decomposition targets~\cite{huang_empirical_1998,luukko_introducing_2016}.

\section{Conclusions, Limitations and Future Work}
\label{sec:conclusion}

This work presented IMFACT, a model-agnostic framework for counterfactual explanation of time series classifiers that operates in the EMD space. 
By decomposing signals into physically meaningful IMFs and substituting them with IMFs drawn from a NUN, IMFACT produces counterfactuals that are valid, proximal, and plausible in temporal structure and value range. 
Across IMF-selection strategies and multi-NUN configurations on two benchmark datasets, nearly all variants achieved high success rates while preserving key signal characteristics; the one exception, \texttt{maxmin} on FaultDetectionA, shows strategy choice can be dataset-dependent rather than universal. 
No single strategy dominates, \texttt{variance} is consistently most plausible, but the most proximal strategy varies by dataset, while cyclic multi-NUN with three neighbours offers the most consistent overall trade-off. 
Against Wachter, Native Guide, and Glacier, IMFACT is the only method to reach full validity on both datasets, with substantially lower and more stable runtime than the strongest competing baselines. 
These findings support signal-decomposition-native perturbation spaces for time series counterfactual explanation, particularly where frequency content and temporal coherence are critical to a plausible explanation.

\textit{Limitations and future work --} 
The evaluation uses sampled subsets of two datasets; full-dataset evaluation across more UCR benchmarks with confidence intervals over multiple seeds is needed to confirm generality. 
Single-sample traces show sensitivity to the specific NUN retrieved, and all experiments use a SimpleCNN, so behaviour on deeper architectures (ResNet, InceptionTime, Transformers) with more complex decision boundaries remains untested; IMFACT is model-agnostic by construction but this has not been verified empirically. 

The method is also currently limited to univariate series, extending to multivariate data requires a principled way to align IMFs across channels. 
Two further limitations concern the core mechanism itself: (i) EMD is prone to mode mixing, so the same IMF index need not correspond to the same physical phenomenon across the query and NUN, and our zero-padding heuristic for $K \neq K'$ does not resolve this; and (ii) the greedy, one-IMF-at-a-time search is not guaranteed to find the globally sparsest or most proximal counterfactual, and the best-performing IMF-selection strategy varies by dataset (e.g.\ \texttt{maxmin} fails outright on FaultDetectionA), so some per-dataset tuning is currently required rather than a single default that works everywhere. 

On the computational side, \texttt{variance} is the slowest strategy despite using the fewest iterations (11--13 vs.\ 61--103 on FruitFlies), since each substitution targets a large, energetically dominant IMF; Ensemble EMD or Variational Mode Decomposition~\cite{dragomiretskiy_variational_2013} could reduce this cost. 
We also observed runtime volatility of up to two orders of magnitude across repeated passes for the baselines, while IMFACT's runtime stayed stable; a more controlled timing protocol is needed before drawing firm conclusions about relative runtime. 
Finally, we plan to add further plausibility metrics (density-based measures, adversarial detectability, human ratings) to more comprehensively assess realism and trustworthiness.

\begin{credits}
\subsubsection{\ackname}
We thank the creators of the UCR/UEA Time Series Archive for enabling rigorous evaluation of these methods. 
We are also grateful to the researchers whose work we reviewed for their contributions to interpretable machine learning. 
\end{credits}

%
%
%
\bibliographystyle{splncs04}
\bibliography{xkdd2026}
\end{document}